\documentclass{article}
\usepackage{graphicx} %
\usepackage{url}
\usepackage{wrapfig}
\usepackage{hyperref}
\PassOptionsToPackage{numbers}{natbib} %
\usepackage[dblblindworkshop, final]{neurips_2026}
\workshoptitle{Resource-Aware Agentic AI}
\usepackage{booktabs}
\usepackage{multirow}
\usepackage[table]{xcolor}
\usepackage{graphicx, subcaption}
\usepackage[export]{adjustbox} %
\usepackage{amsmath}

\definecolor{qgreen}{RGB}{0,100,0}
\definecolor{kred}{RGB}{180,0,0}
\definecolor{oblue}{RGB}{0,70,180}
\definecolor{rewardblue}{RGB}{50,100,200}

\usepackage{xcolor}

\definecolor{timeblue}{RGB}{31,78,121}
\definecolor{scoreorange}{RGB}{180,95,6}

\newcommand{\tval}[1]{\textcolor{timeblue}{#1}}
\newcommand{\sval}[1]{\textcolor{scoreorange}{#1}}
\newcommand{\tbval}[1]{\textcolor{timeblue}{\mathbf{#1}}}

\newcommand{\std}[1]{{\scriptsize $\pm #1$}}

\usepackage{amsmath}
\usepackage[most]{tcolorbox}
\usepackage{cleveref}
\tcbset{
  promptstyle/.style={
    colback=gray!10,
    colframe=black,
    fonttitle=\bfseries,
    boxrule=0.5pt,
    arc=4pt,
    outer arc=4pt,
    boxsep=5pt,
    left=5pt,
    right=5pt,
    top=5pt,
    bottom=5pt,
    enhanced,
  }
}

\title{On the Clock: Towards Punctual and Productive Time-Budgeted AI Agents}

\author{
Aaron Wang\thanks{Work done during an internship at AWS Agentic AI.}\\
University of Illinois at Chicago\\
\And
Neelabh Madan\footnotemark[1]\\
New York University\\
\And
Vlad Sobal\\
AWS\\
\And
Matthew Trager\\
AWS\\
\And
Michael Kleinman\\
AWS\\
\And
Elman Mansimov \\
AWS\\
\And
Wei Xia\\
AWS\\
\And
Stefano Soatto\\
AWS\\
}\date{August 2026}

\begin{document}

\maketitle

\begin{abstract}
We study whether small LLM agents can operate effectively under explicit wall-clock time budgets by both respecting the allocated runtime and using available time productively. We evaluate Qwen3.6-27B on five competitions from MLE-Bench Lite and Qwen3-4B on Zork I (Jericho), two agentic benchmarks where additional computational time can meaningfully improve performance. In the simplest setting, where the budget is stated only in the prompt, agents fail to translate the stated budget into controlled use of time. These failures arise from gaps in time awareness, since the harness provides no timing feedback, but also because they cannot reliably anticipate the duration of actions, and do not have a learned mapping from available time to an appropriate strategy. We investigate two complementary classes of interventions: harness-based mechanisms that expose timing information and enforce deadlines, and reinforcement learning with budget-aware rewards. Injecting timing information through the harness substantially improves budget adherence for Qwen3.6-27B without measurable loss in performance, while enforcement hooks tighten adherence further. RL with GRPO achieves near-perfect budget adherence on Zork I and generalizes to held-out budgets not seen during training, but does not improve task performance over the untrained harness on MLE-Bench. Once agents are made to respect the budget, they still fail to use additional time to improve task performance. RL-trained policies learn when to stop but often fill extra time with repeated actions, and GRPO training on multiple budgets tends to collapse toward the strategy learned for the shortest budget. Our results reveal a gap between time adherence and productive time allocation, which remains a central challenge for budget-conditioned agents.
\end{abstract}

 \section{Introduction}

  \begin{figure}[t]
    \centering
    \includegraphics[width=\linewidth]{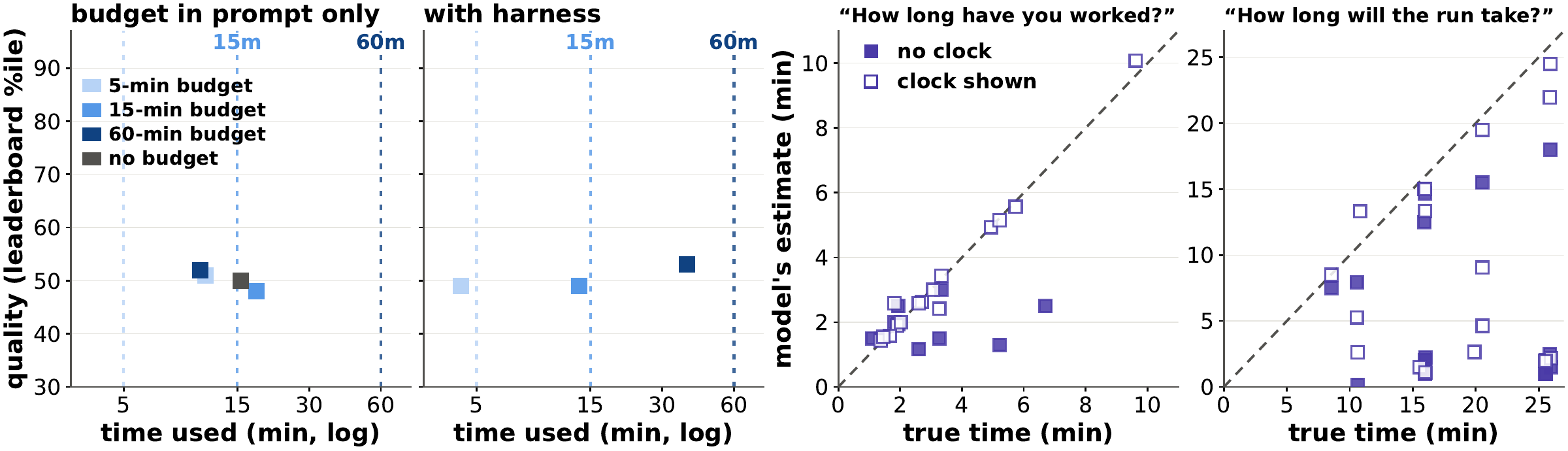}
\caption{\textbf{(Left two)} Mean quality (leaderboard percentile) against
  mean wall-clock time used by Qwen3.6-27B on MLE-Bench, with the budget stated
  in the prompt (left) and with the enforcement harness added (right). Color
  gives the stated budget and dashed lines mark the budget values.
  \textbf{(Right two)} Mid-episode time estimation on 15-minute MLE-Bench
  tasks. The agent is asked how long it has been working and how long the run
  will take in total, and its answer is plotted against the true value, with
  the dashed diagonal marking a perfect estimate. Open markers have a clock in
  context, filled markers do not.}
    \label{fig:teaser}
  \end{figure}

  As AI agents take on complex, long-horizon tasks involving planning, tool use,
  and iterative refinement, how long they spend on a task becomes an important
  dimension of their behavior, alongside the quality of the final solution.
  Agents should be able to operate effectively under explicit wall-clock time
  budgets for two reasons. First, agents face \emph{external constraints}. A
  user should be able to control how long a task takes end to end, and an
  agent embedded in a larger system must meet deadlines to coordinate with
  other components. Second, for many tasks \emph{time is aligned with quality}.
  In search-heavy problems such as ML engineering, more time allows an agent to
  explore more architectures, hyperparameters, or candidate solutions, so
  making agents as fast as possible is not the goal. At the same time, users
  differ in how they weigh solution quality against time, and a
  time budget is how a user expresses that trade-off.

  We require two things from an agent operating under a budget, which we call
  \emph{adherence} and \emph{productive use}. Adherence means the agent returns
  a solution within the allocated time; a harness can force this with a hard
  interrupt, so that the agent works to the deadline itself.
  Productive use means the agent spends that time well, using
  additional budget to improve the solution rather than stopping early or
  padding its runtime with unnecessary actions. A concrete test is whether task performance improves as the budget increases. If the agent achieves the same quality under a tight budget as under a generous one, the additional time is not being converted into a better solution. Much of these tasks are also dominated by tool call latencies, and we test how well the model is able to take in both the latencies and its generation time in order to stay in budget. 

  Current agents fail at adherence even in the simplest setting, where the
  budget is stated in the prompt and the harness gives no feedback on elapsed
  time. On MLE-Bench, Qwen3.6-27B does not use the allocated time effectively (Figure~\ref{fig:teaser}, left; Table~\ref{tab:initial}).
  Qwen3.6-27B runs for roughly ten minutes regardless of the stated budget,
  overrunning an 8-minute budget while using a small fraction of a 60-minute budget.

\begin{wraptable}{r}{0.48\textwidth}
    \centering
    \small
    \setlength{\tabcolsep}{4pt}
    \setlength{\extrarowheight}{1.2pt}

    \resizebox{\linewidth}{!}{%
    \begin{tabular}{llccc}
    \toprule
    Model & Budget & On time & Elapsed & Quality \\
    \midrule
    \multirow{4}{*}{Qwen3.6-27B}
      & \phantom{0}8 min & $35\%$  & $12.9$ \std{12.6} & $50.3 \pm 2.9$ \\
      & 15 min & $83\%$  & $\phantom{0}9.7$ \std{10.6} & $43.1 \pm 8.6$ \\
      & 60 min & $100\%$ & $10.5$ \std{6.3}  & $54.8 \pm 7.2$ \\
      & none   & --      & $12.2$ \std{11.4} & $53.5 \pm 7.8$ \\
    \bottomrule
    \end{tabular}%
    }

    \caption{Experiments on the Dog Breed Identification Task MLE-Bench Lite Task,
    budget stated in the prompt only. Qwen3.6-27B runs $8$--$13$ min regardless
    of the prompt. \emph{On time} is the fraction of runs finishing within the
    stated limit; elapsed is minutes, mean \std{s.d.}; quality is leaderboard
    percentile over valid submissions, mean $\pm$ s.e.m.}
    \label{tab:initial}
\end{wraptable}

  These failures reflect concrete gaps in what the model can observe and
  predict about time. Nothing in the model's context indicates elapsed wall-clock time, and token count is
  a poor proxy since it depends on hardware and inference configuration.
  Probing the models directly confirms this. Asked mid-episode how long it has
  been working, Qwen3.6 27b misjudges its own elapsed time by a large amount
  on a 15-minute task (Figure~\ref{fig:teaser}, right;
  Table~\ref{tab:timecheck}, Appendix). Agents also cannot reliably predict
  how long their actions will take, which matters most for long-running tool
  calls such as training jobs, and even with a clock in context, forecasts of
  the run's remaining and total duration stay far off. We hypothesize that
  even an agent with perfect timing information must still learn to map available time to an appropriate plan.

  We investigate two complementary classes of interventions: harness-based and learning-based.
  \emph{Harness-based} interventions close the observational gaps. A per-action
  clock and periodic reminders report the remaining budget and the time cost
  of each action, and an enforcement mode optionally holds the deadline by
  clamping or refusing over-long commands.
  \emph{Learning-based} interventions target the planning gap. We also explore reinforcement learning (GRPO) against budget-aware rewards as a preliminary step toward internalizing the time--quality trade-off directly. We evaluate on two agentic
  settings where additional time can meaningfully improve outcomes, five competitions from MLE-Bench Lite and Zork I (Jericho),
  which differ in action structure, feedback density, and how time converts into
  quality. We select a
  small number of tasks where different budgets should favor different
  strategies, and analyze failure modes in detail.

  In this work, we separate the problem of time budgeting into adherence and productive use, two parts of very different
  difficulty. We show that \emph{adherence} is largely solvable in our settings. The harness
  brings Qwen3.6-27B from $161\%$ of budget to on-budget with no measurable loss in quality (between the runs with and without the harness), and RL yields near-exact adherence on Zork I that generalizes
  to held-out budgets, though training is less effective on MLE-Bench, where
  the search space is much larger. \emph{Productive use}, however, is much harder. We show that agents that stay on time do not convert additional time into better outcomes, and the RL policy that learns when to stop fills
  extra time with actions that do not produce better task results. 

\section{Related Work}

\paragraph{Budget-awareness and time-aware test-time scaling.}
Lin et al.~\cite{lin2026bagen} formalize budget-awareness as progressive interval estimation, where the agent predicts upper and lower bounds on remaining budget at each step, and find that frontier agents are consistently over-optimistic about how much budget remains. Their SFT+RL training improves early-stop behavior, but precise calibration remains difficult. Our work focuses on wall-clock time rather than token budgets, and studies the progression from prompting through harness enforcement to RL rather than a single training method. Li et al.~\cite{li2026bavt} propose a training-free Budget-Aware Value Tree Search that shifts from exploration to exploitation as budget depletes, and McCleary and Ghawaly~\cite{mccleary2026bcas} measure how search depth, retrieval strategy, and completion budget affect accuracy under fixed constraints. Both operate at the token or tool-call level. We study wall-clock time, whose stochastic action durations token counting cannot capture.

Snell et al.~\cite{snell2024scaling} show that a compute-optimal test-time scaling strategy can outperform a $14\times$ larger model, but assume an unlimited budget. Ma et al.~\cite{ma2026timely} redefine test-time scaling in wall-clock terms with the Timely-Eval benchmark and a Timely-RL training method, and find that existing models fail to adapt reasoning to time budgets. Where Ma et al. focus on a single RL intervention, we compare prompting, harness enforcement, and RL as a progression. Paglieri et al.~\cite{paglieri2025planning} use SFT and RL to teach agents when planning is worthwhile. We study how the entire strategy should adapt to a time budget, and find that RL learns when to stop but not what to do differently with more time.

\paragraph{Budgeted and constrained reinforcement learning.} Constrained Markov Decision Processes~\cite{altman1999constrained} formalize reward maximization subject to cost constraints. Carrara et al.~\cite{carrara2019budgeted} extend this to Budgeted MDPs, conditioning a single policy on the budget so it solves the CMDP for any budget level without retraining, an idea our RL stage adopts. Our cost signal, wall-clock time, is stochastic and unpredictable, while classical BMDPs assume known cost functions of state-action pairs. McMahan and Zhu~\cite{mcmahan2024anytime} require that the budget never be violated at any point in time, closer to our setting since an agent that overruns at step $t$ cannot undo it at $t+1$. Their result that Markovian policies are insufficient under anytime constraints informs our choice to inject remaining budget into the agent's context. Achille~\cite{achille2026universal} argues that time is the critical quantity to optimize when scaling reasoning models, and warns of a failure mode in which reward-driven models brute-force solutions without acquiring transferable strategies. Our RL-trained policy's memorized-prefix-plus-no-ops behavior matches this failure mode.

\paragraph{Reinforcement learning for LLMs.} Group Relative Policy Optimization (GRPO)~\cite{shao2024deepseekmath}, introduced for DeepSeek-Math and used for DeepSeek-R1~\cite{guo2025deepseekr1}, samples a group of rollouts per task and normalizes rewards within the group, avoiding a separate critic network. We adopt GRPO because this normalization handles the multi-modal reward distribution produced by training across several budget levels at once.

\section{Setup}

\subsection{Datasets}

We evaluate time-budgeted agent behavior in two environments: machine learning engineering with MLE-Bench, and text-based games with Jericho Zork. We chose both because task performance scales meaningfully with wall-clock time, they admit multiple valid strategies at different budget levels, and they give graded scores for partially completed work.

\paragraph{MLE-Bench Lite.}

MLE-Bench~\cite{chan2024mlebench} evaluates agents on end-to-end machine learning tasks derived from Kaggle competitions. We use MLE-Bench Lite, a 22-competition subset, and select 5 tasks from it based on two criteria. The task must admit qualitatively different strategies at different time scales (e.g.\ a quick baseline versus iterative feature engineering), and preliminary runs must show non-trivial score variance across time budgets ranging from 10 to 25 minutes. Each task requires the agent to inspect data, develop and execute modeling code, iterate on experiments, and produce a valid submission. The environment is well suited for studying time-aware agents because additional wall-clock time enables different activities. A short budget may suffice for a single baseline model, while a longer budget allows time for data analysis, feature engineering, model selection, hyperparameter tuning, and longer training runs.  We use a five-competition subset of MLE-Bench Lite, namely dog-breed identification and dogs-vs-cats (image classification), leaf classification (species classification from extracted leaf-image features), spooky-author identification (text classification), and NOMAD~2018 transparent-conductor prediction (tabular regression).

We use Qwen3.6-27B~\cite{qwen3} as the underlying language model, run through the OpenCode agentic harness. For RL training, we use an internal tool to obtain TITO (Training In, Training Out) formatted data from agent rollouts, converting trajectories into training samples for GRPO. We vary the wall-clock budget and measure how the agent's trajectory and final task performance change as a function of available time.

\paragraph{Jericho (Zork I).}

Jericho~\cite{hausknecht2020interactive} is an interactive-fiction environment in which an agent interacts with a text-based game through natural-language actions. We use \texttt{zork1} for all experiments. At each step the model observes the current game state, generates an action, and receives the resulting observation. Additional time translates directly into more exploration steps, and performance scales with how much of the world map the agent discovers and how many puzzles it solves along the way.

For evaluations we use Qwen3-4B~\cite{qwen3}, Qwen3.5-9B and Qwen3.8-27B \cite{qwen35blog} in a lightweight agentic loop with thinking mode off. The model repeatedly receives the observation, generates the next action, and observes the result until the trajectory terminates or the time budget is exhausted. We define the task reward, $R_\mathrm{task}$ as the normalized game score,
which provides a dense, comparable measure of progress across trajectories ($\mathrm{max\_score} = 350$ for Zork I).
We use this environment for the Jericho Zork1 RL experiments, training the models to condition its behavior on different time budgets and to trade off exploration, task progress, and budget adherence.

\section{Results}

\subsection{Prompting}

  \begin{table}[t]
    \centering
    \small
    \begin{tabular}{lccccc}
    \toprule
    Condition & On budget (\%) & Quality (\%ile) & Turns & Time (s) &
  Valid (\%) \\
    \midrule
    Prompt only                  & 15 & $49.4 \pm 4.4$ & $22.4 \pm 10.1$ & $1453
  \pm 454$ & 78 \\
    \quad + Harness (clocks)     & 62 & $42.1 \pm 1.8$ & $16.5 \pm 6.4$  & $897
  \pm 202$  & 88 \\
    \quad + Enforcement (clamp)  & 74 & $46.8 \pm 1.8$ & $17.0 \pm 6.5$  & $847
  \pm 165$  & 90 \\
    \quad + Enforcement (refuse) & 88 & $42.4 \pm 1.9$ & $20.3 \pm 8.0$  & $812
  \pm 160$  & 78 \\
    \quad + RL (GRPO, refuse)    & 86 & --             & $18.9 \pm 3.7$  & $807
  \pm 37$   & 74 \\
    \bottomrule
    \end{tabular}
    \caption{\textbf{MLE-Bench results}. We show incremental effects of in-context clocks, budget enforcement, and RL at a fixed 15-minute budget (900 seconds), all conditions sharing an
  identical prompt (RL
    row explained in Section~\ref{sec:rl-training}). \emph{On budget} is the
  fraction of runs
    finishing within the stated limit; quality is leaderboard percentile over
    submissions that are both valid and on budget, mean $\pm$ s.e.m.; turns and
    time are mean $\pm$ s.d. Pooled over five MLE-bench competitions.}
    \label{tab:ablation}
  \end{table}

\paragraph{MLE-Bench.} Prompting alone produced weak adherence to the specified time budget. Without direct access to wall-clock time, models cannot reliably estimate how long previous actions, tool calls, or reasoning steps have taken. A model could in principle track time itself by querying a system clock, but we did not observe this. The total wall-clock time consumed was largely insensitive to the budget stated in the prompt: on MLE-Bench with prompting only, the agent used 161 percent of a 15-minute budget on average, and only 15 percent of runs finished within the stated limit (\Cref{tab:ablation}).

\paragraph{Zork I.}

\begin{figure}[t]
    \centering
    \includegraphics[width=\linewidth]{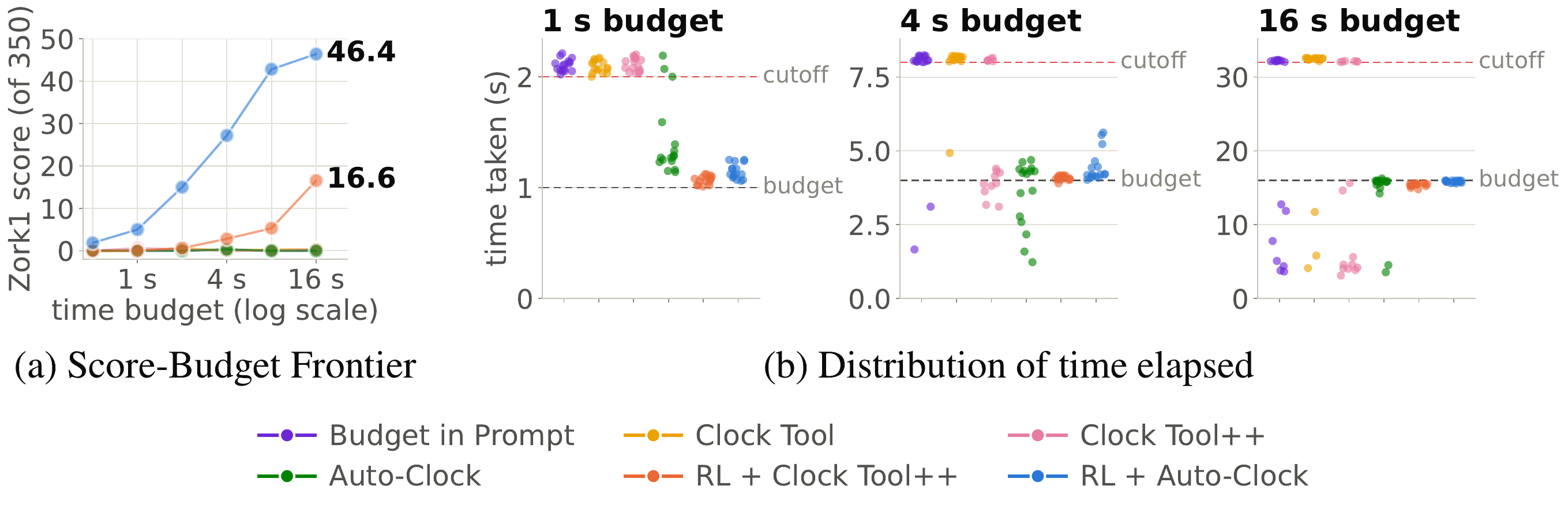}
    \caption{
    Zork1 performance and time expenditure for Qwen3-4B with \textbf{thinking mode disabled} under per-episode time budgets, in four fixed-weight prompting conditions and two policies trained for 400 GRPO steps against the budget ($n=16$ episodes per cell; episodes are terminated at $2\times$ the budget, which serves as a hard cutoff). (a) Mean score against budget: the trained policies attain 46.4 and 16.6 points at 16 s, whereas no fixed-weight condition exceeds 0.6 points at any budget. (b) Elapsed time for the same episodes at budgets of 1, 4, and 16 s, with one marker per episode and horizontal jitter; dashed gray denotes the budget and dashed red the $2\times$ budget hard cutoff. Stating the budget in the prompt alone does not meaningfully improve adherence, while AutoClock is the most effective harness-based intervention. RL yields the strongest overall performance, achieving both the best budget adherence and substantially higher task scores.
    }
    \label{fig:zork_figure}
\end{figure}

Stating the time budget in the prompt alone does not lead to reliable budget adherence. As shown by the \textbf{\texttt{Budget in Prompt}} rows in Table~\ref{tab:clock_tool_time_score_nothink}, all models substantially overshoot the stated budget, often running until the hard cutoff at $2\times$ the budget terminates the episode (See Fig \ref{fig:zork_figure}.b). Task performance also remains near zero for most models. This behavior is expected: the models have no direct notion of wall-clock generation speed and receive no feedback about how much time has elapsed during execution. As a result, a stated budget provides a constraint in natural language, however out of the box models do not exhibit the ability to apply the constraint.

\subsection{Interventions via Harness}

\paragraph{MLE-Bench.} Exposing elapsed time through the harness made the agent substantially more responsive to its budget. Three  hooks (Table~\ref{tab:hooks}) report the time remaining and the cost of generation, without prescribing any action. This reduced the time consumed from $161.4\%$ to $99.7\%$ of the stated budget, with no measurable change in quality ($47.5$ to $47.0$ percentile, within error), and raised the fraction of runs producing a valid submission from $78\%$ to $88\%$ (Table~\ref{tab:ablation}).

\begin{table}[t]
\centering
\small
\begin{tabular}{llp{0.38\linewidth}}
\toprule
\textbf{Hook} & \textbf{When it fires} & \textbf{What it does} \\
\midrule
Per-tool clock & After every tool call & Adds the time remaining to the result the agent reads \\
Periodic reminder & On a fixed time interval & Reports the time remaining even while a command is still running \\
Generation accounting & After every tool call & Reports how long the model spent generating text before the action \\
\midrule
Enforcement (clamp) & Before a shell command runs & Shortens the command's time limit so it stops by the deadline; the command still runs, but is cut off \\
Enforcement (refuse) & Before a shell command runs & Blocks the command and returns a message explaining why, so the agent must choose something cheaper \\
\bottomrule
\end{tabular}
\caption{Harness hooks. The first three only add information to what the agent reads. The two enforcement modes change the action itself and are never used together.}
\label{tab:hooks}
\end{table}

The dominant failure mode was that when the agent recognized that little time remained, it launched a tool call whose duration exceeded that budget (for example a training  with a timeout that is longer than the time remaining). Because the duration of a tool call is variable and hard to predict, the agent often had no further chance to act before it ran past the budget. 
Even when prompted to put tool calls in the background so it can interrupt the runs, the model is still unable to consistently stop in time. 

We therefore added an enforcement hook that inspects each shell command before it runs, comparing the time limit the agent set for the command against the time remaining. Because commands are frequently issued with no limit, a missing limit is treated as the harness default timeout. Enforcement cut the overrun further, at a small cost in quality. The on-time rate improved, while quality fell slightly.   The quality cost is unsurprising, since the unenforced model was converting overrun time into score, and enforcement takes that time away. Shortening a command's limit preserved most of the unenforced quality and produced the highest rate of valid submissions we measured ($90\%$). Blocking commands outright held closer to the budget but reduced the valid formatted submission rate while increaseing interaction steps, because the agent reissued blocked commands with smaller limits.

\begin{figure}[t]
  \centering
  \includegraphics[width=\linewidth]{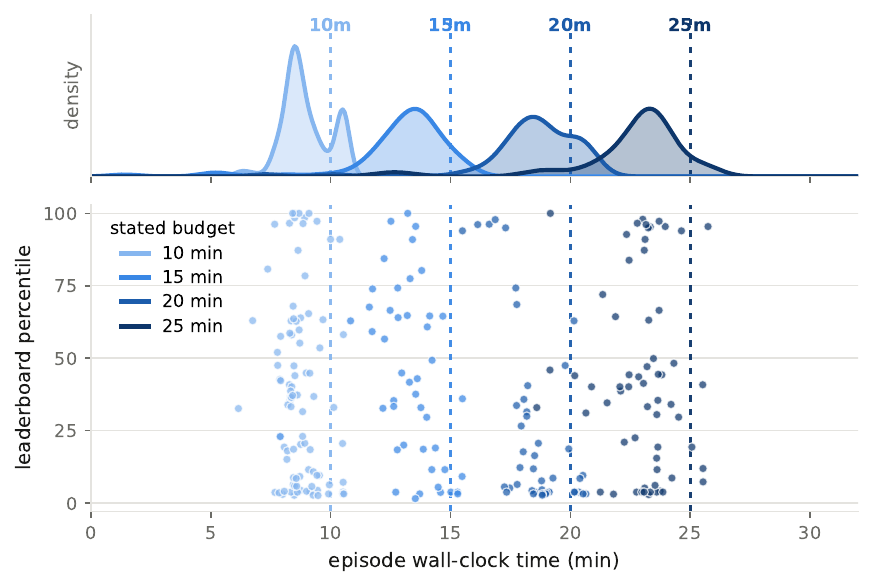}
  \caption{Time and quality distributions on MLE-Bench under the harness (clocks and enforcement), without RL training. Top panel shows elapsed-time density per budget. Bottom panel shows leaderboard percentile against elapsed time for valid submissions. Dotted vertical lines mark each stated budget. The agent scales time to the budget, but quality does not improve with additional time.}
  \label{fig:time-dist-untrained}
\end{figure}

Even without RL training, the agent scales its elapsed time to the stated budget when given a clock. Figure~\ref{fig:time-dist-untrained} (Appendix) shows the time and quality distributions across four budgets. Quality is spread across the full percentile range at every budget, with no upward trend as budget grows.

\paragraph{Zork I.}

To provide agents with an explicit sense of elapsed time, we introduce three harness-level interventions in Zork. First, the \texttt{Clock Tool} allows the model to issue a \texttt{time\_elapsed} tool call, which returns a message of the form \texttt{[time used: $x$s of $B$s budget]}. Second, \texttt{Clock Tool++} uses the same tool interface but provides substantially more guidance in the prompt on when and how to query it, together with an in-context example demonstrating its use. Third, the \texttt{Auto-Clock} harness removes the need for the model to actively request timing information after every interaction turn by having the harness automatically appends the same elapsed-time message to the context.

The basic \texttt{Clock Tool} practically yields no improvements in adherence (See Fig \ref{fig:zork_figure} and Table \ref{tab:clock_tool_time_score_nothink}). Even though the model now has access to wall-clock time, it must still decide when to query the tool, and models often fail to call it frequently enough for the feedback to be useful. Consequently, most models still continue to substantially overshoot short budgets. With \texttt{Clock Tool++}, however, models invoke the tool much more often (Table~\ref{tab:tool-calls-thinking-off} in Appendix), and the resulting increase in timing feedback leads to a comparatively better budget adherence. The strongest adherence comes from the \texttt{Auto-Clock} harness, where timing information is pushed into the context after every turn rather than being requested by the model. Under this intervention, most models track the available budget substantially more closely. However, a limitation is that harness feedback is only available between generations: if a single generation itself consumes a large fraction of the budget, the harness cannot intervene until that generation completes. This effect is particularly visible for larger models such as 9B and 27B. Another interesting observation is that the score remains below $5$ across all budgets, despite substantially improved budget adherence. This suggests that access to time feedback helps the model determine \emph{when to stop}, but does not teach it \emph{how to adapt its strategy}. Therefore, we observe no meaningful score--budget trade-off in the frontier shown in Fig.~\ref{fig:zork_figure}(a).

\subsection{Interventions via RL Training}
\label{sec:rl-training}

\subsubsection{MLE-Bench}
\label{sec:mlebench-rl}

\paragraph{Objective.} We train with GRPO~\cite{shao2024deepseekmath}. Let $B$ be the budget, $t$ the elapsed wall-clock time, and $q \in [0,1]$ the normalized task score of the submission left behind. The reward is
\begin{equation}
R(q,t) =
\begin{cases}
q + c_{\mathrm{time}}, & \text{submitted and } t \le B, \\[4pt]
\alpha\, q - p\,(t - B), & \text{submitted and } t > B, \\[4pt]
-\,c_{\mathrm{fail}}, & \text{no submission},
\end{cases}
\end{equation}
with $c_{\mathrm{time}} > 0$ as a bonus for finishing within the budget, $\alpha \in (0,1)$ as the share of quality kept when late, $p > 0$ as the penalty per second of overrun, and $c_{\mathrm{fail}} > 0$ as the cost of producing nothing. The constants are set so that, at equal quality, finishing on time always beats finishing late, and producing nothing is always worst.

\paragraph{Results.} We show results on MLE-bench in \ref{tab:ablation}. Training did not improve over the harness it was trained under.  Training did reduce interaction steps from $20.3$ to $18.9$, improving upon the retry behavior that the enforcing plugin introduced. 

The trained policy scaled its effort with the budget. The percent of budget used regardless of the set budget held between $0.87$ and $0.93$ in every run. On the other hand, about a quarter of on-time submissions scored no better than a uniform guess, and this fraction did not fall when the weight on $q$ (quality) was tripled. The GRPO training was unable to improve the models ability to perform on the subset of MLEBench tasks, but did enforce its ability to stay on budget. 
\subsubsection{Zork I}

\paragraph{Objective.} Let $B$ be the budget in seconds, stated verbatim in the prompt (see \nameref{sec:Prompts}), $t$ the wall-clock time the episode consumed, and $s = \mathrm{score}/\mathrm{score}_{\max} \in [0,1]$ the normalized game score the agent holds when the episode ends ($\mathrm{score}_{\max} = 350$ for Zork I). The game score is read off the live game directly. The reward must supply a reason to \emph{stop}, since the agent may end the episode itself with \texttt{quit} and otherwise plays until a hard $180$\,s wall-clock kill or the turn cap. Below, we describe the reward functions we explore. All reward functions share the following form:
\begin{equation}
R(s,t) \;=\; \bigl[\, s + \lambda\, b(t; B) \,\bigr] \cdot g(t; B),
\end{equation}
with $b \in [0,1]$ as a time bonus, $g \in \{0,1\}$ as a gate that voids the whole reward, and $\lambda > 0$ weighting time against score (the score always carries weight $1$; we use $\lambda = 0.1$). The four schemes, detailed in Table~\ref{tab:zork1-rewards-appendix} (Appendix), differ only in $b$ and $g$. (i) \emph{Cutoff} is a pure gate with no bonus (ii) \emph{Linear} and (iii) \emph{Gaussian} are smooth peaks at $t = B$ that pay for arriving near the budget from either side (iv) \emph{Step} is a coarse ramp that pays for elapsed time in tenths of $B$ and then voids the episode past $1.1B$. We select the \textit{step-wise} reward for training our RL models. Although the shaped rewards exhibit broadly comparable behavior, \textit{step-wise} provides the best overall balance between convergence speed, task score, and budget adherence (Table~\ref{tab:reward-shaping}). We train the RL policies by sampling a time budget uniformly from 0 to 16s at each GRPO step and optimizing with the \textit{step-wise} reward. We apply this training procedure to both the \texttt{AutoClock} and \texttt{Clock Tool++} interventions and referring the resulting runs as \texttt{RL + AutoClock} and \texttt{RL + Clock Tool++}, respectively. In both settings, the policies learn to remain within the allotted budget while also improving task performance under that budget and pushing the frontier (See Figure \ref{fig:zork_figure}(a)). However, \texttt{RL + Clock Tool++} underperforms \texttt{RL + AutoClock} in score (Fig.~\ref{fig:zork_figure} and Table~\ref{tab:clock_tool_time_score_nothink}). We attribute this gap to the overhead of active time checking since \texttt{RL + Clock Tool++} spends roughly 50\% of its turns querying the clock, leaving fewer actions for task progress. Despite this inefficiency, both policies achieve strong adherence, with generations typically terminating within $\pm 15\%$ of the specified budget.

\begin{figure*}[h]
    \centering

    \begin{subfigure}[t]{0.49\textwidth}
      \centering
      \includegraphics[width=\linewidth,valign=t]{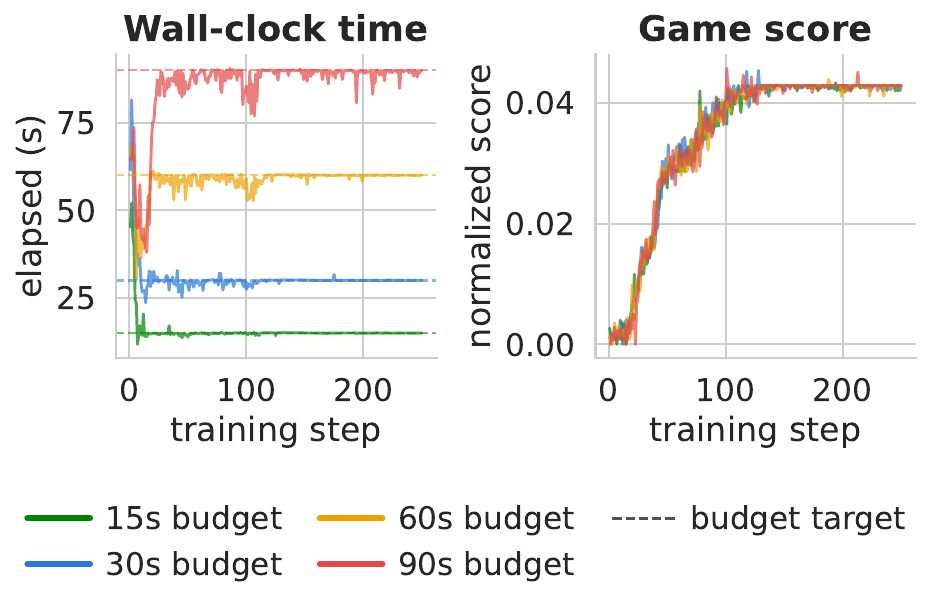}
      \caption{Budget mix $\{15, 30, 60, 90\}$.}
      \label{fig:mixbudget-train-4}
    \end{subfigure}
    \hfill
    \begin{subfigure}[t]{0.49\textwidth}
      \centering
      \includegraphics[width=\linewidth,valign=t]{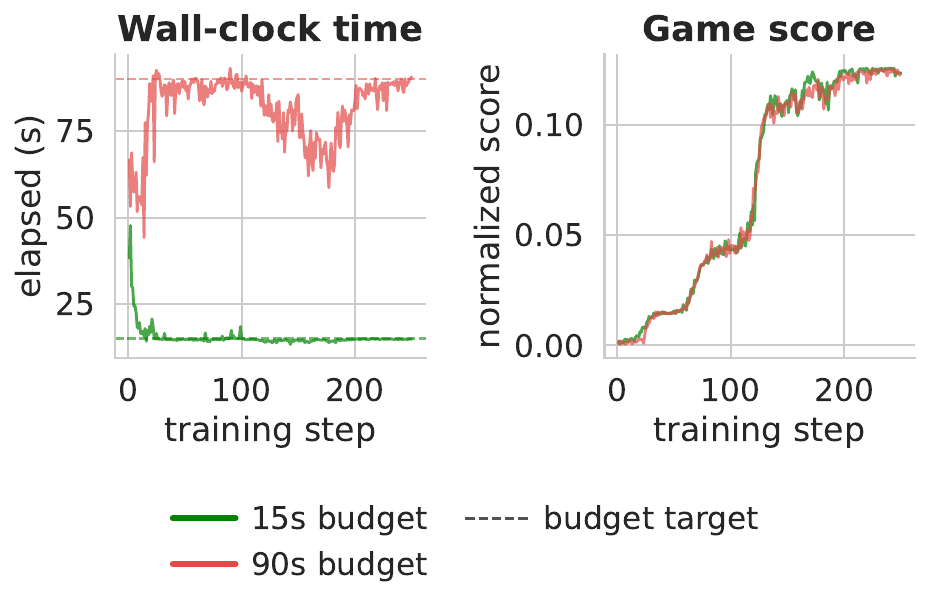}
      \caption{Budget mix $\{15, 90\}$.}
      \label{fig:mixbudget-train-2}
    \end{subfigure}

    \caption{Mixed-budget GRPO training on Zork~I with the step-wise time-shaped reward and
    the per-turn clock shown in the prompt ($\lambda = 0.1$). 
    }
    \label{fig:mixbudget-training}
\end{figure*}

\textbf{Clock generalizes, the play doesn't:} We also train the same policy using discrete budgets rather than sampling uniformly from a continuous range, which allows us to study substantially longer horizons such as 15, 30, 60, and 90,s. Each rollout is assigned a budget from a predefined set we call budget mix, and the corresponding budget is provided to the model in the prompt. 
In principle, interleaving multiple budgets during training should encourage the policy to condition its behavior on the available time and use larger budgets for more extensive exploration. As training progresses, budget adherence becomes nearly perfect across the different horizons. However, the policy collapses toward the strategy learned for the smallest budget (Figure~\ref{fig:mixbudget-training}).  Trajectories for different budgets separate cleanly in elapsed wall-clock time but their game scores follow nearly identical trends throughout training. Thus, the model learns \emph{when to stop} for each requested budget, but not \emph{how to use additional time productively} to improve task performance.

\section{Limitations}

\textbf{Strategy collapse across budgets}: The trained policy learns to closely follow itself to the stated budget but does not plan differently for it. In Zork, rollouts under different budgets open with the same memorized action sequence and reach the same score regardless of budget given. The budget changes only how long the episode is padded with filler actions before the agent quits. Mixed-budget training was intended to prevent this collapse, and it succeeded only at helping the model generalize stopping to multiple budgets. The same pattern holds on MLE-Bench, where extra time bought more attempts training models, but did not improve the final score. Also see \nameref{sec:invalid}.

\textbf{Memorization in fixed environments}: 
Training on a single game with a fixed starting state makes memorization the cheapest path to higher score. The Zork policy converges on a recalled walkthrough prefix that collects a fixed number of points, after which no new score is found. When valid actions are fed back to the model instead of its own outputs, the same checkpoint scores far lower, confirming that the learned behavior does not survive even a small change. The reward structure compounds this problem by pricing elapsed time and not exploration, so once a scoring prefix exists, deviating from it offers no benefit. It also does not distinguish a rejected action from a productive one, and the trained policy fills remaining time by repeating a command. Budget utilization therefore only demonstrates the ability of the model to stop on time, but not on how well it used the allotted time. We discuss these mechanisms, and a third source of noise from stochastic action durations on MLE-Bench, in the Appendix.

\textbf{Scope of evaluation}: 
Our experiments use two models (Qwen3-4B and Qwen3.6-27B) and two environments, and the RL experiments on Zork train on a single game. Whether the findings extend to frontier-scale models, to environments with richer strategy spaces remains open. The MLE-Bench subset contains five competitions, which limits the diversity of task structures the policy encounters during training. 
A broader evaluation across models, environment types, and task counts would strengthen the conclusions drawn here.

\section{Conclusion}

We studied whether LLM agents can reason about and act under explicit time budgets, comparing two classes of interventions: (i) harness-based mechanisms that expose or enforce time information, and (ii) GRPO training that directly optimizes budget-conditioned behavior. Simply stating a time budget in the prompt is not sufficient: out-of-the-box agents struggle to estimate action durations, overshooting small budgets by up to $2\times$ on Zork and by $61\%$ on MLE-Bench. In contrast, exposing time through the agent harness substantially improves adherence. Informational clocks reduce overruns to near zero in both environments with little measurable loss in task quality, while stricter enforcement hooks tighten adherence further at a modest performance cost.

GRPO training improves budget adherence further. On Zork, the trained policy matches the requested budget to within $0.2$\,s across budgets from $5$ to $90$\,s, including four held-out budgets not seen during training. On MLE-Bench, trained policies consistently use between $0.87$ and $0.93$ of the allocated budget. At smaller budgets, RL also pushes the cost--score frontier outward, improving both adherence and task performance. However, this improvement does not extend uniformly to larger budgets. In these sparse-reward regimes, the current objectives provide little incentive for deeper exploration or qualitatively different strategies. On Zork, the policy converges to a fixed high-reward prefix and spends additional time on unproductive actions, causing score to saturate at $45/350$ across larger budgets. Similarly on MLE-Bench, additional time similarly leads to more attempts at essentially the same training procedure without improving submission quality.

We show that these results reveal a distinction between \emph{budget adherence} and \emph{productive budget allocation} (ability to change its behavior when given more time). These findings should be interpreted in light of three limitations: strategy collapse across budgets, memorization induced by fixed training environments and sparse rewards, and the limited scope of our evaluation across models and tasks. Possible directions include exploration-aware rewards that pay for novel states rather than elapsed time, environment variation that prevents memorization from substituting for search, and reward terms that distinguish repeated attempts from genuinely different strategies. The gap between adherence (which RL solves) and productive time allocation (which remains open) defines the central challenge for budget-conditioned agents.

\newpage

\bibliographystyle{unsrt}
\bibliography{bibliography}

\newpage
\section*{Appendix}

\subsection*{Reward structure and exploration incentives}

The reward is a function of final score and elapsed time only, so exploring is unpriced. In Zork, once a scoring prefix exists, deviation is actively risky. Our 4B models typically reach at most about $45/350 \approx 0.129$ in score. With $\lambda = 0.1$, the time term is worth up to $0.100$, which is already comparable to the score term achieved by the policy. Moreover, because Zork rewards are relatively discrete, a meaningful improvement often requires gaining several additional points; gaining five more points, for example, increases the score term by only $5/350 \approx 0.014$. Perfect budget adherence is therefore worth about seven times as much as a genuine five-point discovery, and it is available deterministically from any token sequence that consumes the clock. Fixing this would require a term that pays for novelty (first-visit bonuses, walkthrough coverage, or potential-based shaping on distance to unreached states), together with environment variation across episodes so that recall cannot substitute for search.

On MLE-Bench, the analogous problem is that the reward does not distinguish between repeated attempts at the same strategy and genuinely different approaches. A policy that submits the same baseline three times under a longer budget receives the same reward as one that tries progressively more sophisticated methods, as long as the best submission score is identical.

\subsection*{Budget adherence through invalid actions}
\label{sec:invalid}

Because the Zork time term is satisfied by wall-clock alone, the policy learns to consume the clock by emitting commands the parser rejects. In one particular training at step 500 the filler is the word \texttt{forward}, to which Zork replies \texttt{I don't know the word "forward".} It appears in $127$ of $128$ rollouts, immediately after the memorized prefix ends, and accounts for $45\%$ of turns at $B = 15$ and $88\%$ of turns at $B = 90$. The budget is met to within $0.2$\,s almost entirely through repetition of an action that advances nothing, since elapsed time measures wall-clock latency regardless of whether the action was accepted, and a rejected command consumes the same generation time as a valid move.

This means ``used $99\%$ of its budget'' should not be read as $99\%$ of the budget spent working. Any claim about effort allocation requires a measure of useful time (turns that change game state, valid-command rate, or tokens spent on accepted actions). This failure mode is also easy to miss with the wrong diagnostic. Our repetition detector reports a trailing identical run of only $1.0$--$1.5$ turns for these runs, because it measures the tail identical to the \emph{final} action (\texttt{quit}), so hundreds of consecutive no-ops sit outside its window. Penalizing invalid actions or crediting only state-changing turns is a prerequisite for interpreting time-shaped rewards as effort.

\subsection*{Comparing Reward Shapes}
Across the shaped variants, \emph{linear}, \emph{Gaussian}, and \emph{step} perform comparably overall, suggesting that the precise shape of the time reward matters little once an explicit timing signal is present. The main difference is in budget matching. The \emph{step} reward produces the most reliable adherence to the target budget, since it does not reward the model at all for being over budget. A more detailed comparison of the reward shapes and their resulting behavior is provided in Table~\ref{tab:zork1-rewards-appendix} (Appendix).

\subsection*{Action durations on MLE-Bench}

On MLE-Bench, action durations are stochastic and depend on factors outside the agent's control (GPU availability, data loading, network latency). The same training script may take $30$ seconds or $3$ minutes depending on system load. This makes the reward noisy in a way that is uncorrelated with the agent's decisions, complicating credit assignment during RL~\cite{mcmahan2024anytime}. The harness-only condition already achieves $99.7\%$ budget utilization, leaving little room for RL to improve, and the variance in action duration may explain why RL training on MLE-Bench produced negligible gains over the untrained harness.

\section*{Implementation Details}

\subsection*{Environment and Interaction Protocol}

Each turn the model receives the game's response, a status line
\texttt{[score: $s$/350 | moves: $m$]}, and the fixed question
\texttt{What is your next action?}. It must reply with exactly two lines, a
\texttt{REASONING:} line and an \texttt{ACTION:} line (the full template is in the base
system prompt, reproduced under \emph{Prompts} below). Parsing takes the last line beginning with \texttt{ACTION:};
failing that, the last non-empty line; failing that, the fallback command \texttt{look}.
The extracted command is stripped of surrounding quotes and backticks and truncated to
$190$ characters (Jericho truncates commands at $198$ bytes). A truncated generation
therefore never aborts an episode: it silently degrades to \texttt{look}, and the rate of
such turns is recorded per configuration.

An episode terminates on any of: the agent emitting \texttt{ACTION: quit} (the prompt
requires this, so voluntary termination is the intended exit), game over or victory,
\texttt{max\_turns}~$=4000$, the wall-clock cutoff defined below, a
server-side context-length overflow (recorded separately as \texttt{context\_limit}), or an
LLM error. The end reason is recorded for every episode; the split between
\emph{agent quit} and \emph{harness cutoff} is one of the primary dependent variables, since
it identifies who actually enforced the deadline.

In the benchmark harness the message history is trimmed to the most recent $100$ turn-pairs,
always retaining the system prompt and the opening observation and always resuming on an
assistant message. At the cutoffs used here ($\leq 80$\,s) no episode reaches $100$ turns, so
the trimming is a no-op; it is kept so that longer budgets cannot silently measure context
exhaustion instead of play (an untrimmed episode dies on the context limit at roughly turn
$250$, worst for Qwen3-4B, whose native context is $40{,}960$ tokens).

\subsection*{Time Budget, Cutoff, and the Definition of Elapsed Time}

Every timed episode is given a budget $B$ seconds and is killed by the harness at a cutoff of
$2B$. Elapsed time is wall clock measured from the start of the episode; it includes
generation latency, environment stepping and all overhead, and it is exactly the quantity the
model is shown or can query. Two properties of the cutoff matter for interpreting results,
and both are shared by the benchmark and the training loop by construction:

\begin{itemize}
  \item \textbf{The cutoff is checked before each generation and cannot interrupt one.} Every
        episode therefore receives at least one generation, and elapsed time can exceed the
        cutoff. At a $32$\,s cutoff this is a rounding error; at a $1$\,s cutoff the single
        generation \emph{is} the episode. We report
        $\mathrm{overrun} = \text{elapsed}/\text{cutoff}$ per configuration and mark any cell
        with $\mathrm{overrun} \geq 1.5$ as one in which the cutoff never bound, so that its
        score is not read as timed play. Note that this ratio is against the cutoff, not the
        budget: because the cutoff is $2B$, an episode that never stops voluntarily scores
        $\approx 2.0\times$ the budget, which is the ceiling rather than a pathology.
  \item \textbf{Sub-second budgets are below the latency of a single turn.} One generation
        costs $\approx 0.2$--$0.4$\,s for the 4B model at 16-way concurrency and $1.4$--$1.8$\,s
        for the 27B; the fastest complete episode observed anywhere is $0.233$\,s. Zork~I's
        first point is $\approx 12$ moves in. Budgets below $\approx 2$\,s are consequently at
        the score floor by construction, and we use them only to measure clock behaviour
        (tool calls, quit rate, elapsed time), never as score conditions.
\end{itemize}

Elapsed time is rendered to the model at a precision the budget can resolve: whole seconds for
$B \geq 10$\,s, one decimal for $1 \leq B < 10$\,s, two decimals below $1$\,s. Without this, a
$0.5$\,s budget would render every turn as \texttt{[time used: 0s of 0s budget]} and erase the
quantity under study.

\subsection*{The Clock Interventions}

The four interventions differ \emph{only} in how the time budget reaches the model. All four state
the same budget in the system prompt and use the same $2B$ cutoff, the same game, the same
seeds and the same sampling parameters. Table~\ref{tab:impl-conditions} summarises them; the
exact prompt text of each is given under \emph{Prompts} at the end of this section.
A fifth condition,
\emph{No Budget}, states nothing about time and serves as the control.

\begin{table}[t]
  \centering
  \small
  \begin{tabular}{lccc r}
    \toprule
    Condition & Budget in & Per-turn & Callable & System prompt \\
              & prompt    & clock    & clock    & at $B{=}16$\,s \\
    \midrule
    No Budget        & --  & --  & --  & 1301 \\
    Budget in Prompt & yes & --  & --  & 1456 \\
    Clock Tool       & yes & --  & yes & 1766 \\
    Clock Tool++     & yes & --  & yes & 3013 \\
    Auto-Clock       & yes & yes & --  & 1456 \\
    \bottomrule
  \end{tabular}

  \caption{The clock conditions. \emph{Per-turn clock} appends
  \texttt{[time used: $x$s of $B$s budget]} to every observation; \emph{callable clock} lets the
  model spend a turn on \texttt{ACTION: time\_elapsed} to obtain the same string.
  \emph{Clock Tool} and \emph{Clock Tool++} expose the identical tool and return the identical
  reply, and differ only in how it is documented. \emph{Budget in Prompt} and
  \emph{Auto-Clock} send the identical system prompt and differ only in the user turns. The
  last column is the system-prompt length in characters.}
  \label{tab:impl-conditions}
\end{table}

\paragraph{The clock tool is a text-protocol pseudo-tool, not native function calling.}
The model requests the clock by replying \texttt{ACTION: time\_elapsed} in place of a game
command, and receives the elapsed/budget line back without the game advancing: no move is
consumed, the score is untouched, and the game state is unchanged. The only cost is the real time the
generation itself took, which is charged to the same budget. This was chosen over OpenAI-style
\texttt{tools=[\ldots]} deliberately: native tool calling would route the three model families
through different vLLM tool-call parsers and a different chat-template branch than the
non-tool conditions use, adding a confound to the single comparison the experiment exists to
make. The cost of the choice is that the condition measures whether a model \emph{chooses to
check the clock}, not whether it can emit a tool-call schema.

\subsection*{Reinforcement Learning}

\paragraph{Algorithm and loop.} We train with GRPO in verl, using a custom agent loop that
plays Jericho inside the rollout worker. Generated tokens carry
$\texttt{response\_mask} = 1$; game observations, the clock-tool replies and every other
environment-authored user turn carry $0$ and are excluded from the policy-gradient loss. The
loop is registered as the default agent loop and Jericho is stepped through an executor, so
the synchronous environment does not block the asynchronous rollout. The trajectory reward is
computed by the loop at episode end and returned directly, bypassing the reward manager.

\paragraph{Reward.} With $s$ the normalised game score (score$/350$), $t$ the episode's elapsed
wall clock and $B$ its budget, the reward is
\begin{equation}
  R = s + \lambda \cdot \mathrm{bonus}(t, B), \qquad \lambda = 0.1,
\end{equation}
with the \emph{step-wise} bonus used in every run reported here:
\begin{equation}
  \mathrm{bonus}(t,B) =
  \begin{cases}
    \min\!\left(1,\; \tfrac{1}{10}\left(\left\lfloor \frac{t}{0.1B} \right\rfloor + 1\right)\right)
      & t \leq 1.1B,\\[4pt]
    0 \quad \text{and } R := 0 & t > 1.1B .
  \end{cases}
\end{equation}
The bonus rises in ten steps of $0.1$ across the budget, saturates at $1.0$ on the
$0.2B$-wide plateau $[0.9B, 1.1B]$ --- the $+1$ in the tenth index means the plateau begins at
$0.9B$, not at $B$ --- and past $1.1B$ the \emph{entire} reward including the score term is
gated to zero. Two consequences matter for reading the results. The risk is asymmetric:
overshooting costs every point in the game while undershooting costs only partial credit. And
because the plateau is wide and everything outside it is gated, the mean bonus over a set of
trajectories is in practice a \emph{pass rate} for ``landed in $[0.9B, 1.1B]$'' rather than a
graded distance from $B$; we therefore report $t/B$ alongside it, which is continuous and is
recorded even for gated episodes. Three further schemes are implemented and available but not
used in the reported runs: a linear tent ($0$ at $t=0$, $1$ at $t=B$, $0$ at $t \geq 2B$), a
Gaussian centred on $B$ with $\sigma = 10$\,s, and a hard gate ($s$ if $t \leq B$, else $0$).

\paragraph{Hyperparameters.} Table~\ref{tab:impl-rl} lists the settings, which are unchanged
from our Zork~I sweeps so that the runs stay comparable.

\begin{table}[t]
  \centering
  \small
  \begin{tabular}{ll}
    \toprule
    Algorithm & GRPO (\texttt{adv\_estimator=grpo}), KL loss on, \texttt{kl\_loss\_coef} $=0.001$ \\
    Policy & Qwen3-4B, trained from the base model (no warm start) \\
    Learning rate & $1 \times 10^{-6}$ \\
    Batching & train batch $16$, PPO mini-batch $16$, micro-batch $1$ per GPU, \\
             & gradient checkpointing on \\
    Group size & $n = 8$ rollouts per prompt $\Rightarrow 128$ episodes per step \\
    Sampling & temperature $1.0$, thinking disabled \\
    Context & prompt $\leq 4096$, response $\leq 32768$, \texttt{max\_model\_len} $40960$ \\
             & (Qwen3-4B native; no RoPE scaling) \\
    Episode & \texttt{max\_turns} $4000$, \texttt{max\_assistant\_turns} $4000$, \\
            & \texttt{max\_action\_tokens} $2048$, kill $= 2 \times B$ \\
    Rollout & vLLM, asynchronous mode, TP${=}1$, \texttt{gpu\_memory\_utilization} $0.5$, \\
            & $16$ agent workers \\
    Data & $512$ train rows, $16$ validation rows, no valid-action list \\
    Schedule & $20$ epochs $= 640$ steps; checkpoint every $50$ steps; validate every $10$ \\
    Hardware & $1$ node $\times$ $8$ H200 per run \\
    \bottomrule
  \end{tabular}
  \caption{GRPO configuration, identical across runs}
  \label{tab:impl-rl}
\end{table}

\subsection*{Prompts}
\label{sec:Prompts}

\begin{tcolorbox}[title={Base system prompt (all conditions)}, promptstyle, fontupper=\small, breakable]
\begin{verbatim}
You are an expert player of text-based adventure games. Your goal is to
make as much progress as possible and maximize your score. Points are
awarded for discovering new locations, solving puzzles, and
collecting/using items.

Each turn you are shown the current game text. Respond with a single
short command that the game's parser can understand. Common commands
include: look, inventory, movement directions (north, south, east, west,
up, down, enter, exit), examine X, open X, take X, drop X, read X, push
X, turn on X, put X in Y, give X to Y, attack X with Y. Any object X or
Y must be a SINGLE WORD the game recognizes.

Guidelines:
- Explore systematically; examine objects and rooms mentioned in
  descriptions.
- Do NOT repeat a command that just failed; try something different.
- Do not use the 'help', 'save', or 'restore' commands.
- You MUST end the session yourself by responding with ACTION: quit once
  you have finished the game or believe you cannot make further
  progress. Ending with quit is required to complete the session.

Your response MUST follow this exact format and contain nothing else:
REASONING: [1-2 short sentences explaining your choice]
ACTION: [the single short command to execute]

Example:
REASONING: The mailbox might contain something useful.
ACTION: open mailbox
\end{verbatim}
\end{tcolorbox}

The budget statement is appended to the base prompt in every condition except
\emph{No Budget}, which is the control and is told nothing about time.

\begin{tcolorbox}[title={Budget statement}, promptstyle, fontupper=\small]
\begin{verbatim}
IMPORTANT: You have a time budget of 16 seconds to finish the game. Act
efficiently and aim to make as much progress as possible within this
time budget.
\end{verbatim}
\end{tcolorbox}

The two tool blocks document the same tool, which returns the same reply; they differ only in
how heavily it is advertised. 

\begin{tcolorbox}[title={\emph{Clock Tool}: brief tool documentation}, promptstyle, fontupper=\small]
\begin{verbatim}
You cannot see the clock, but you may check it. Responding with
ACTION: time_elapsed
instead of a game command returns how much of your time budget you have
used so far. This does NOT advance the game and does not count as a
move, but the check itself takes real time out of your budget, so use it
sparingly.
\end{verbatim}
\end{tcolorbox}

The detailed version bundles three changes relative to the brief one (i) full mechanics, (ii) a
worked example turn, (iii) an instruction to check early and act on the answer.

\begin{tcolorbox}[title={\emph{Clock Tool++}: detailed tool documentation}, promptstyle, fontupper=\small, breakable]
\begin{verbatim}
CHECKING THE CLOCK
You cannot see the clock, but you have a tool that reports it.

How to call it: reply with
ACTION: time_elapsed
in place of a game command. You get back a line of the form [time used:
3s of 16s budget].

What it costs: the game does not advance - no move is used, the score
does not change, and the room stays exactly as it was. The only cost is
the real time the reply itself takes, which comes out of the same
budget.

Use it, and start using it EARLY. Check within your first few actions,
before you have committed to a plan. An early check tells you how fast
your turns are running, which lets you work out roughly how many more
actions the remaining budget buys and pace yourself accordingly; check
again occasionally to stay calibrated. A check made at the very end
tells you nothing you can still act on, so early checks are worth far
more than late ones. Then act on what it tells you: if the budget is
going faster than you expected, stop exploring and go straight for
progress, and once most of it is gone, wrap up and respond with ACTION:
quit.

Worked example of a clock check mid-session:

West of House
There is a small mailbox here.

[score: 0/350 | moves: 3]

What is your next action?

REASONING: Before exploring further I should see how much of my budget I
have spent.
ACTION: time_elapsed

[time used: 3s of 16s budget]
(Clock check only - the game did not advance.)

What is your next action?

REASONING: Only about a fifth of the budget is gone after 3 moves, so I
can afford to keep exploring north.
ACTION: north
\end{verbatim}
\end{tcolorbox}

The first user turn carries the game's opening text and the status line. The clock line is
present under \emph{Auto-Clock} only, where it starts at zero.

\begin{tcolorbox}[title={First user turn}, promptstyle, fontupper=\small]
\begin{verbatim}
You are now playing the game. Here is the opening:

<game opening text>

[score: 0/350 | moves: 0]
[time used: 0s of 16s budget]

What is your next action?
\end{verbatim}
\end{tcolorbox}

Subsequent turns are identical in structure, with the game's response to the last command in
place of the opening.

\begin{tcolorbox}[title={Subsequent user turns}, promptstyle, fontupper=\small]
\begin{verbatim}
<game response to the last command>

[score: 5/350 | moves: 12]
[time used: 7s of 16s budget]

What is your next action?
\end{verbatim}
\end{tcolorbox}

A clock check is answered without advancing the game. The reply reports elapsed and total,
never the difference: the model has to subtract. Its first line is byte-identical to the line
\emph{Auto-Clock} pushes unbidden, which is what isolates pull-versus-push as the only
difference between those conditions.

\begin{tcolorbox}[title={Reply to \texttt{ACTION: time\_elapsed}}, promptstyle, fontupper=\small]
\begin{verbatim}
[time used: 3s of 16s budget]
(Clock check only - the game did not advance.)

What is your next action?
\end{verbatim}
\end{tcolorbox}

\begin{table*}[t]
\centering
\small

\resizebox{\textwidth}{!}{%
\begin{tabular}{lcccccc@{\hspace{8pt}}cccccc}
\toprule
& \multicolumn{6}{c}{\textbf{\textcolor{timeblue}{Elapsed Time (s)}}}
& \multicolumn{6}{c}{\textbf{\textcolor{scoreorange}{Score (max = 350)}}} \\
\cmidrule(lr){2-7}
\cmidrule(lr){8-13}

\textbf{Budget (s) $\rightarrow$}
& \textbf{0.5} & \textbf{1} & \textbf{2} & \textbf{4} & \textbf{8} & \textbf{16}
& \textbf{0.5} & \textbf{1} & \textbf{2} & \textbf{4} & \textbf{8} & \textbf{16} \\
\midrule

\multicolumn{13}{l}{\textbf{Qwen3-4B}} \\

Budget in Prompt
& \tval{$1.1{\scriptstyle\pm0.0}$}
& \tval{$2.1{\scriptstyle\pm0.1}$}
& \tval{$4.1{\scriptstyle\pm0.1}$}
& \tval{$7.4{\scriptstyle\pm2.0}$}
& \tval{$15.8{\scriptstyle\pm1.0}$}
& \tval{$21.2{\scriptstyle\pm13.1}$}
& \sval{$0.0{\scriptstyle\pm0.0}$}
& \sval{$0.0{\scriptstyle\pm0.0}$}
& \sval{$0.0{\scriptstyle\pm0.0}$}
& \sval{$0.3{\scriptstyle\pm1.2}$}
& \sval{$0.0{\scriptstyle\pm0.0}$}
& \sval{$0.0{\scriptstyle\pm0.0}$} \\

\quad + \texttt{Clock Tool}
& \tval{$1.1{\scriptstyle\pm0.0}$}
& \tval{$2.1{\scriptstyle\pm0.1}$}
& \tval{$4.0{\scriptstyle\pm0.2}$}
& \tval{$7.9{\scriptstyle\pm0.8}$}
& \tval{$15.1{\scriptstyle\pm2.8}$}
& \tval{$27.7{\scriptstyle\pm10.3}$}
& \sval{$0.0{\scriptstyle\pm0.0}$}
& \sval{$0.0{\scriptstyle\pm0.0}$}
& \sval{$0.6{\scriptstyle\pm2.5}$}
& \sval{$0.0{\scriptstyle\pm0.0}$}
& \sval{$0.3{\scriptstyle\pm1.2}$}
& \sval{$0.3{\scriptstyle\pm1.2}$} \\

\quad + \texttt{Clock Tool++}
& \tval{$1.1{\scriptstyle\pm0.1}$}
& \tval{$2.1{\scriptstyle\pm0.1}$}
& \tval{$3.1{\scriptstyle\pm0.7}$}
& \tval{$5.4{\scriptstyle\pm2.1}$}
& \tval{$11.5{\scriptstyle\pm4.4}$}
& $\tbval{16.1{\scriptstyle\pm13.3}}$
& \sval{$0.0{\scriptstyle\pm0.0}$}
& \sval{$0.6{\scriptstyle\pm2.5}$}
& \sval{$0.3{\scriptstyle\pm1.2}$}
& \sval{$0.0{\scriptstyle\pm0.0}$}
& \sval{$0.0{\scriptstyle\pm0.0}$}
& \sval{$0.3{\scriptstyle\pm1.2}$} \\

\quad + \texttt{Auto-Clock}
& \tval{$1.1{\scriptstyle\pm0.1}$}
& \tval{$1.4{\scriptstyle\pm0.3}$}
& $\tbval{2.3{\scriptstyle\pm0.3}}$
& $\tbval{3.6{\scriptstyle\pm1.1}}$
& $\tbval{8.2{\scriptstyle\pm2.2}}$
& $\tbval{14.2{\scriptstyle\pm4.0}}$
& \sval{$0.0{\scriptstyle\pm0.0}$}
& \sval{$0.0{\scriptstyle\pm0.0}$}
& \sval{$0.0{\scriptstyle\pm0.0}$}
& \sval{$0.3{\scriptstyle\pm1.2}$}
& \sval{$0.0{\scriptstyle\pm0.0}$}
& \sval{$0.0{\scriptstyle\pm0.0}$} \\

\midrule
\multicolumn{13}{l}{\textbf{Qwen3-4B + Time-Budget RL (GRPO, 400 steps)}} \\

\quad RL + \texttt{Clock Tool++}
& \tval{$0.8{\scriptstyle\pm0.2}$}
& $\tbval{1.1{\scriptstyle\pm0.0}}$
& $\tbval{2.0{\scriptstyle\pm0.1}}$
& $\tbval{4.1{\scriptstyle\pm0.1}}$
& $\tbval{8.1{\scriptstyle\pm0.1}}$
& $\tbval{15.4{\scriptstyle\pm0.2}}$
& \sval{$0.0{\scriptstyle\pm0.0}$}
& \sval{$0.0{\scriptstyle\pm0.0}$}
& \sval{$0.6{\scriptstyle\pm1.7}$}
& \sval{$2.8{\scriptstyle\pm2.6}$}
& \sval{$5.3{\scriptstyle\pm5.3}$}
& \sval{$16.6{\scriptstyle\pm10.9}$} \\

\quad RL + \texttt{Auto-Clock}
& \tval{$0.8{\scriptstyle\pm0.2}$}
& $\tbval{1.1{\scriptstyle\pm0.1}}$
& $\tbval{2.2{\scriptstyle\pm0.1}}$
& $\tbval{4.5{\scriptstyle\pm0.5}}$
& $\tbval{8.2{\scriptstyle\pm0.1}}$
& $\tbval{15.8{\scriptstyle\pm0.1}}$
& \sval{$1.9{\scriptstyle\pm4.0}$}
& \sval{$5.0{\scriptstyle\pm0.0}$}
& \sval{$15.0{\scriptstyle\pm0.0}$}
& \sval{$27.2{\scriptstyle\pm14.4}$}
& \sval{$42.8{\scriptstyle\pm7.5}$}
& \sval{$46.4{\scriptstyle\pm2.1}$} \\

\midrule
\multicolumn{13}{l}{\textbf{Qwen3.5-9B}} \\

Budget in Prompt
& \tval{$1.2{\scriptstyle\pm0.1}$}
& \tval{$3.0{\scriptstyle\pm2.7}$}
& \tval{$4.4{\scriptstyle\pm0.5}$}
& \tval{$10.3{\scriptstyle\pm3.2}$}
& \tval{$17.8{\scriptstyle\pm3.3}$}
& \tval{$30.1{\scriptstyle\pm10.2}$}
& \sval{$0.0{\scriptstyle\pm0.0}$}
& \sval{$0.0{\scriptstyle\pm0.0}$}
& \sval{$0.0{\scriptstyle\pm0.0}$}
& \sval{$0.9{\scriptstyle\pm2.7}$}
& \sval{$0.6{\scriptstyle\pm1.7}$}
& \sval{$2.2{\scriptstyle\pm3.6}$} \\

\quad + \texttt{Clock Tool}
& \tval{$1.2{\scriptstyle\pm0.1}$}
& \tval{$2.2{\scriptstyle\pm0.1}$}
& \tval{$4.5{\scriptstyle\pm0.9}$}
& \tval{$9.4{\scriptstyle\pm2.2}$}
& \tval{$19.7{\scriptstyle\pm4.5}$}
& \tval{$32.6{\scriptstyle\pm6.8}$}
& \sval{$0.0{\scriptstyle\pm0.0}$}
& \sval{$0.0{\scriptstyle\pm0.0}$}
& \sval{$0.0{\scriptstyle\pm0.0}$}
& \sval{$0.3{\scriptstyle\pm1.2}$}
& \sval{$2.2{\scriptstyle\pm3.1}$}
& \sval{$2.2{\scriptstyle\pm3.6}$} \\

\quad + \texttt{Clock Tool++}
& \tval{$1.4{\scriptstyle\pm0.4}$}
& \tval{$2.1{\scriptstyle\pm0.1}$}
& \tval{$4.3{\scriptstyle\pm0.3}$}
& \tval{$9.0{\scriptstyle\pm1.1}$}
& \tval{$17.9{\scriptstyle\pm2.6}$}
& \tval{$29.5{\scriptstyle\pm7.1}$}
& \sval{$0.0{\scriptstyle\pm0.0}$}
& \sval{$0.0{\scriptstyle\pm0.0}$}
& \sval{$0.0{\scriptstyle\pm0.0}$}
& \sval{$0.3{\scriptstyle\pm1.2}$}
& \sval{$1.6{\scriptstyle\pm3.5}$}
& \sval{$2.5{\scriptstyle\pm3.2}$} \\

\quad + \texttt{Auto-Clock}
& \tval{$1.3{\scriptstyle\pm0.4}$}
& \tval{$3.5{\scriptstyle\pm3.6}$}
& \tval{$4.6{\scriptstyle\pm2.8}$}
& \tval{$9.0{\scriptstyle\pm2.3}$}
& \tval{$14.4{\scriptstyle\pm4.0}$}
& \tval{$22.1{\scriptstyle\pm6.8}$}
& \sval{$0.0{\scriptstyle\pm0.0}$}
& \sval{$0.0{\scriptstyle\pm0.0}$}
& \sval{$0.0{\scriptstyle\pm0.0}$}
& \sval{$0.0{\scriptstyle\pm0.0}$}
& \sval{$1.2{\scriptstyle\pm3.4}$}
& \sval{$2.5{\scriptstyle\pm4.1}$} \\

\midrule
\multicolumn{13}{l}{\textbf{Qwen3.8-27B}} \\

Budget in Prompt
& \tval{$1.1{\scriptstyle\pm0.1}$}
& \tval{$2.3{\scriptstyle\pm0.2}$}
& \tval{$4.4{\scriptstyle\pm0.2}$}
& \tval{$8.5{\scriptstyle\pm0.2}$}
& \tval{$16.5{\scriptstyle\pm0.2}$}
& \tval{$32.4{\scriptstyle\pm0.1}$}
& \sval{$0.0{\scriptstyle\pm0.0}$}
& \sval{$0.0{\scriptstyle\pm0.0}$}
& \sval{$0.0{\scriptstyle\pm0.0}$}
& \sval{$0.6{\scriptstyle\pm2.5}$}
& \sval{$1.6{\scriptstyle\pm3.0}$}
& \sval{$5.3{\scriptstyle\pm4.3}$} \\

\quad + \texttt{Clock Tool}
& \tval{$1.1{\scriptstyle\pm0.1}$}
& \tval{$2.2{\scriptstyle\pm0.1}$}
& \tval{$4.1{\scriptstyle\pm0.4}$}
& \tval{$8.3{\scriptstyle\pm0.3}$}
& \tval{$16.6{\scriptstyle\pm0.2}$}
& \tval{$31.7{\scriptstyle\pm2.4}$}
& \sval{$0.0{\scriptstyle\pm0.0}$}
& \sval{$0.0{\scriptstyle\pm0.0}$}
& \sval{$0.0{\scriptstyle\pm0.0}$}
& \sval{$0.0{\scriptstyle\pm0.0}$}
& \sval{$2.2{\scriptstyle\pm4.1}$}
& \sval{$3.4{\scriptstyle\pm3.0}$} \\

\quad + \texttt{Clock Tool++}
& \tval{$1.5{\scriptstyle\pm0.1}$}
& \tval{$2.3{\scriptstyle\pm0.2}$}
& \tval{$3.4{\scriptstyle\pm0.5}$}
& \tval{$8.1{\scriptstyle\pm0.4}$}
& \tval{$12.6{\scriptstyle\pm2.6}$}
& \tval{$21.5{\scriptstyle\pm4.0}$}
& \sval{$0.0{\scriptstyle\pm0.0}$}
& \sval{$0.0{\scriptstyle\pm0.0}$}
& \sval{$0.0{\scriptstyle\pm0.0}$}
& \sval{$0.0{\scriptstyle\pm0.0}$}
& \sval{$0.6{\scriptstyle\pm1.7}$}
& \sval{$1.9{\scriptstyle\pm3.1}$} \\

\quad + \texttt{Auto-Clock}
& \tval{$1.3{\scriptstyle\pm0.2}$}
& \tval{$2.2{\scriptstyle\pm0.1}$}
& \tval{$2.9{\scriptstyle\pm0.4}$}
& $\tbval{4.3{\scriptstyle\pm0.4}}$
& \tval{$9.7{\scriptstyle\pm1.9}$}
& $\tbval{16.6{\scriptstyle\pm1.1}}$
& \sval{$0.0{\scriptstyle\pm0.0}$}
& \sval{$0.0{\scriptstyle\pm0.0}$}
& \sval{$0.0{\scriptstyle\pm0.0}$}
& \sval{$0.0{\scriptstyle\pm0.0}$}
& \sval{$0.0{\scriptstyle\pm0.0}$}
& \sval{$2.8{\scriptstyle\pm3.6}$} \\

\bottomrule
\end{tabular}%
}

\caption{
Elapsed time and raw \textit{Zork1} score under different time budgets with thinking disabled.
The left block reports elapsed time in seconds, while the right block reports task score.
Values are mean $\pm$ standard deviation over $n=16$ episodes.
\textbf{Elapsed-time entries whose mean is within 15\% of the stated budget are shown in bold}.
All prompting conditions use the same stated budget and cutoff ($2\times$ budget), differing only in how clock information is communicated.
The Qwen3-4B GRPO rows use time-budget RL trained for 400 steps.
}
\label{tab:clock_tool_time_score_nothink}
\end{table*}

\begin{table}[t]
\centering
\small
\begin{tabular}{lcccc}
\toprule
\textbf{Reward} & \textbf{On-target (\%)} & \textbf{Score (of 350)}
& \textbf{Median $t/B$} & \textbf{Steps to score $\geq 10$} \\
\midrule
\texttt{linear}       & \textbf{100.0} & 8.7           & 1.00 & -- \\
\texttt{gaussian}     & 99.7           & 20.8          & 1.00 & 89 \\
\texttt{step\_wise}   & 96.1           & \textbf{22.5} & 1.00 & \textbf{56} \\
\bottomrule
\end{tabular}

\caption{Comparison of shaped reward formulations for time-budgeted RL on \textsc{Zork1}. 
All three rewards achieve strong budget adherence, with median elapsed time matching the
specified budget. Among them, \texttt{step\_wise} achieves the highest task score and reaches
a score of at least 10 substantially earlier in training than \texttt{gaussian}.}
\label{tab:reward-shaping}
\end{table}

\begin{table}[t]
\centering
\small
\begin{tabular}{llcccccc}
\toprule
setting & model & 0.5\,s & 1\,s & 2\,s & 4\,s & 8\,s & 16\,s \\
\midrule
\multicolumn{8}{l}{\emph{thinking off}} \\
\addlinespace[1pt]
\texttt{Clock Tool}   & Qwen3-4B    & 0.00 & 0.00 & 0.19 & 0.00 & 0.81 & 0.00 \\
\texttt{Clock Tool}   & Qwen3.5-9B  & 0.00 & 0.00 & 0.00 & 0.00 & 0.00 & 0.00 \\
\texttt{Clock Tool}   & Qwen3.8-27B & 0.25 & 0.19 & 0.25 & 0.12 & 0.06 & 0.38 \\
\addlinespace[2pt]
\texttt{Clock Tool++} & Qwen3-4B    & 0.19 & 0.81 & 1.75 & 2.06 & 2.75 & 3.12 \\
\texttt{Clock Tool++} & Qwen3.5-9B  & 0.88 & 0.50 & 0.50 & 0.19 & 0.31 & 1.00 \\
\texttt{Clock Tool++} & Qwen3.8-27B & 1.00 & 0.94 & 0.94 & 1.00 & 1.81 & 2.12 \\
\bottomrule
\end{tabular}

\caption{Clock-tool usage with thinking disabled. Each cell is the mean number of
\texttt{ACTION: time\_elapsed} calls per episode over 16 episodes of \textsc{zork1}.
\texttt{Clock Tool} and \texttt{Clock Tool++} expose the same clock tool and differ only
in how the tool is documented in the system prompt.}
\label{tab:tool-calls-thinking-off}
\end{table}

\begin{table}[t]
    \centering
    \small
    \setlength{\tabcolsep}{5pt}
    \begin{tabular}{llrrrr}
    \toprule
     & & \multicolumn{4}{c}{Mean signed error \std{s.d.} (s)} \\
    \cmidrule(lr){3-6}
    Model & Clock & Elapsed & Budget left & Time to finish & Total runtime \\
    \midrule
    \multirow{2}{*}{Qwen3.6-27B}
      & no  & $-69$ \std{110} & $+43$ \std{381} & $-25$ \std{366} & $-735$
  \std{546} \\
      & yes & $+46$ \std{187} & $-19$ \std{58}  & $-53$ \std{214} & $-554$
  \std{538} \\
    \bottomrule
    \end{tabular}
    \caption{Mid-episode time-estimation probes on MLE-Bench (15-minute budget),
    the data behind Figure~\ref{fig:teaser} (right). The agent is asked how much
    time has elapsed, how much budget remains, how much longer it will run, and
    what the total run duration will be. Errors are prediction minus ground
    truth in seconds. Qwen3.6-27B is probed by
  truncating
    its own recorded episodes, with the clock condition set by keeping or
    stripping the harness timer notices ($n{=}9$--$20$ per cell; $54$--$67$ of
    $80$ probes parsed, the rest ignored the question and kept emitting tool
    calls). With a clock in context both models read elapsed time and remaining
    budget accurately. Without one, even elapsed time is off
    by more than a minute on average. Forecasting the run is hard in every
    condition, and Qwen3.6-27B underestimates its total runtime by roughly ten
    minutes with or without the clock. Reading the clock does not imply
    anticipating how long work will take.}
    \label{tab:timecheck}
  \end{table}

\begin{table*}[t]
    \centering
    \small
    \setlength{\tabcolsep}{5pt}
    \begin{tabular}{@{}p{0.50\linewidth} m{30mm} p{0.23\linewidth}@{}}
      \toprule
      \textbf{Reward} & \centering\textbf{Shape} & \textbf{Behaviour} \\
      \midrule

      \textcolor{rewardblue}{\textbf{\texttt{Unshaped}}}\\[-2pt]
      $R = s$
      & \includegraphics[width=\linewidth]{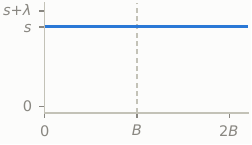}
      & Flat in $t$, just the game score as reward \\
      \addlinespace

      \textcolor{rewardblue}{\textbf{\texttt{Cutoff}}}\\[-2pt]
      $R = s \cdot \mathbf{1}[\,t \le B\,]$
      & \includegraphics[width=\linewidth]{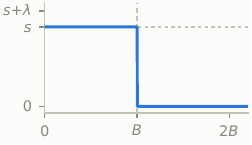}
      & Meet the deadline or forfeit the score. Finishing much earlier than the budget
        receives the same reward as finishing exactly at the budget. \\
      \addlinespace

      \textcolor{rewardblue}{\textbf{\texttt{Linear}}}\\[-2pt]
      $R = s + \lambda
      \max\!\left(0,\,1-\dfrac{|t-B|}{B}\right)$
      & \includegraphics[width=\linewidth]{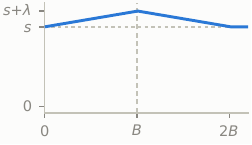}
      & The time bonus rises linearly to its maximum at $t=B$ and then decreases linearly,
        reaching zero at $t=2B$. \\
      \addlinespace

      \textcolor{rewardblue}{\textbf{\texttt{Gaussian}}}\\[-2pt]
      $R = s + \lambda
      \exp\!\left(-\dfrac{(t-B)^2}{2\sigma^2}\right)$
      & \includegraphics[width=\linewidth]{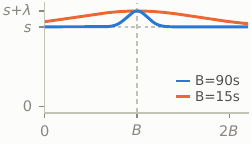}
      & The bonus is largest at $t=B$ and decreases smoothly away from the target.
        We use $\sigma = 10\,\mathrm{s}$. \\
      \addlinespace

      \textcolor{rewardblue}{\textbf{\texttt{Step}}}\\[-2pt]
      $R =
      \left[
      s + \lambda
      \min\!\left(1,\dfrac{\lfloor 10t/B \rfloor + 1}{10}\right)
      \right]
      \mathbf{1}[\,t \le 1.1B\,]$
      & \includegraphics[width=\linewidth]{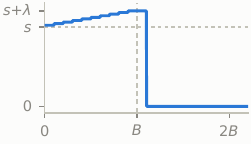}
      & The time bonus increases in discrete tenths as more of the budget is used, with
        the entire reward forfeited after $1.1B$. \\

      \bottomrule
    \end{tabular}

    \caption{Time-shaped rewards for Zork~I. Here $s \in [0,1]$ is the normalized game
    score, $t$ is the episode wall-clock time, $B$ is the time budget, and
    $\lambda = 0.1$ controls the weight of the time bonus. For the Gaussian reward,
    $\sigma = 10\,\mathrm{s}$.}
    \label{tab:zork1-rewards-appendix}
\end{table*}

  \section*{Implementation Details: MLE-Bench}

  \subsection*{Agent, Model Serving, and Sandbox}

  The agent is OpenCode 1.17.4, run headless (\texttt{opencode run --format
  json}) with tool
  calls auto-approved. It uses OpenCode's built-in tools (\texttt{bash},
  \texttt{read},
  \texttt{write}, \texttt{edit}, \texttt{glob}, \texttt{grep}) and OpenCode's
  own system
  prompt; the task, paths, and budget are the first user message (reproduced
  under
  \emph{Prompts}). The agent is never given a clock it can call. In every
  condition it either
  reads the clock lines the harness appends to tool results or has no clock.

  Qwen3.6-27B is served by SGLang with tensor parallelism $4$ and a
  $131{,}072$-token context.
  OpenCode is configured with the same context limit and an output limit of
  $16{,}384$ tokens
  per generation, and compacts its history as it nears the limit. A proxy
  between OpenCode and
  the inference server records the token ids and log-probabilities of every
  generation in an
  episode, with a loss mask that is $1$ on model-generated tokens and $0$ on
  everything the
  environment authored, including tool results and the clock lines inside
  them. These records
  are the training data.

  Each episode runs in its own Docker container with one dedicated NVIDIA
  GPU, one episode per
  container at a time, in a fresh workspace that is deleted afterwards. The
  competition's public
  files are linked into the workspace read-only. The private test labels sit
  on the same
  read-only mount and are read by the grader; the prompt tells the agent not
  to read them, and
  nothing else prevents it. After every episode, all processes it started are
  killed before
  grading, so a background job cannot change \texttt{submission.csv} after
  the deadline or slow
  the next episode.

  \subsection*{Episode, Budget, and Elapsed Time}

  An episode is one OpenCode invocation. It ends when the agent exits or when
  the harness kills
  it. In the Table~\ref{tab:ablation} conditions the kill is at $2B$
  ($1800$\,s for the
  $15$-minute budget); the RL run later moved it to $B + 30$\,s (see
  \emph{Reinforcement
  Learning}). whatever
  \texttt{submission.csv} is on disk at that moment is graded.

  Elapsed time $t$ is wall clock from the moment the workspace is ready to
  the moment the agent
  exits or is killed. It includes generation, tool execution, and harness
  overhead, and it is
  the quantity the clock hooks display. The start time is stored in a
  root-owned file the agent
  cannot write. The reward takes the larger of the container's reading and
  the generator's own
  clock, and a failed reading resolves to the cutoff rather than to zero, so
  no failure can
  grade as an on-time finish.

  \emph{On budget} means $t \le B$. \emph{Turns} is the number of assistant
  steps in the
  trajectory, one step being one generation. \emph{Valid} is the fraction of
  episodes whose
  file grades as a valid submission.

  \subsection*{Grading and Quality}

  After the episode, \texttt{mlebench.grade.grade\_csv} grades
  \texttt{submission.csv} against
  the private test labels inside the container and returns whether the file
  is a valid
  submission (ids, columns, row count, no missing values) and the competition
  metric. All
  competitions in the tables are lower-is-better: multi-class log-loss for
  dog-breed
  identification, spooky-author identification, and leaf classification;
  log-loss for
  dogs-vs-cats; mean column-wise RMSLE for NOMAD~2018.

  \emph{Quality} is the leaderboard percentile: the percentage of entries on
  the competition's
  Kaggle leaderboard, as distributed with MLE-Bench, that the episode's score
  beats, ties
  counted as half. It is a monotone transform of the metric and puts the
  competitions on one
  scale.

  \subsection*{Table~\ref{tab:initial} and Figure~\ref{fig:teaser} (left)}

  The budget-in-prompt runs on dog-breed identification predate the harness
  above. They were
  made with an earlier launcher around the same OpenCode agent and model,
  with no plugins, no
  resume loop, and no kill: an episode ended when the agent exited. Each
  budget has its own
  prompt file; the four files share one body of operating instructions and
  differ only in the
  budget sentences reproduced under \emph{Prompts}. The no-budget prompt
  omits the time
  sentences entirely.

  \subsection*{Reinforcement Learning}

  \paragraph{Reward.} The reward in Section~\ref{sec:mlebench-rl} is a
  simplified statement of
  the function the trainer received. Let $x = t/B$ and $L$ the graded metric.
  The quality term
  is
  \begin{equation}
    q = c_q \cdot \frac{L_{\mathrm{bad}} - L}{L_{\mathrm{bad}} -
  L_{\mathrm{good}}},
  \end{equation}
  where $L_{\mathrm{bad}}$ is the competition's uniform-guess metric ($\ln$
  of the number of
  classes for the log-loss tasks; the 90th percentile of observed scores for
  NOMAD, which has
  no analytic chance level) and $L_{\mathrm{good}}$ is the 5th percentile of
  scores observed in
  earlier runs. Chance scores $q = 0$, so a copy of
  \texttt{sample\_submission.csv} or a
  constant prediction, which grades as valid, earns no quality credit. The
  5th-percentile score
  earns $q = c_q$. Below chance the slope is one fifth of the in-band slope
  and $q$ is floored
  at $-1.5$, so miscalibrated submissions stay ordered but one outlier cannot
  dominate a group;
  above $L_{\mathrm{good}}$ the line continues unchanged. Let $Q = q +
  c_{\mathrm{sub}}$, with
  $c_{\mathrm{sub}}$ a flat bonus for any gradeable file. Then
  \begin{equation}
    R =
    \begin{cases}
      Q + c_{\mathrm{time}} + s\,Q\,(1 - x), & \text{valid and } t \le B,
  \\[3pt]
      \kappa\,Q - p_{\max}\bigl(1 - e^{-(t - B)/\tau}\bigr), & \text{valid
  and } t > B, \\[3pt]
      -\,c_{\mathrm{fail}} - k\,x, & \text{no valid submission.}
    \end{cases}
  \end{equation}
  The speed term is scaled by $Q$ so a poor model submitted quickly does not
  out-score a good
  model on time. The late penalty is strictly decreasing in lateness, with no
  flat region, and
  bounded by $p_{\max}$ so the worst late valid submission still beats the
  best invalid one.
  The invalid penalty grows with $x$ so running to the cutoff with no file is
  worse than
  exiting early with no file. Table~\ref{tab:impl-mle-reward} lists the
  constants. With the
  final set, the worst on-time valid reward exceeds the best late valid
  reward on every
  competition, which we check before each run.

  \begin{table}[t]
    \centering
    \small
    \begin{tabular}{llrr}
      \toprule
      Symbol & Meaning & Steps 1--105 & Steps 106--182 \\
      \midrule
      $q$                & quality term                         & $1/(1+L)$ &
  band, $c_q = 3.0$ \\
      $c_{\mathrm{sub}}$ & bonus for a gradeable file           & $0.25$  &
  $0.25$  \\
      $c_{\mathrm{time}}$& bonus for finishing on budget         & $1.0$   &
  $0.5$   \\
      $s$                & on-budget speed slope                & $0.5$   &
  $0.15$  \\
      $\kappa$           & share of $Q$ kept when late          & $0.3$   &
  $0.8$   \\
      $p_{\max}$         & maximum late penalty                 & $1.35$  &
  $1.35$  \\
      $\tau$             & late-penalty time constant           & $200$\,s &
  $20$\,s \\
      $c_{\mathrm{fail}}$& penalty for no valid submission      & $1.5$   &
  $3.0$   \\
      $k$                & slope of the invalid penalty in $x$  & $1.5$   &
  $1.5$   \\
      \bottomrule
    \end{tabular}
    \caption{Reward constants for the MLE-Bench GRPO run. The run launched
  with the left
    column and the compressive quality map $1/(1+L)$; the band and the right
  column replaced
    them at step $106$, after the kill moved to $B + 30$\,s and lateness no
  longer had to be
    priced. The RL row of Table~\ref{tab:ablation} is from the second phase.}
    \label{tab:impl-mle-reward}
  \end{table}

  \paragraph{Groups and budgets.} One GRPO group is one (competition, budget)
  pair, so
  advantages are computed among episodes told the same deadline. The budget
  in the prompt, the
  budget the plugins display, the kill, and every reward branch are set from
  the same per-group
  value. The run sweeps $5$ competitions $\times$ $4$ budgets ($10, 15, 20,
  25$ minutes): $20$
  groups of $21$ episodes, $420$ episodes per optimizer step, one step per
  sweep. Dogs-vs-cats
  is the fifth competition, included so that every step has $20$ groups, and
  is excluded from
  the tables. All episodes in a step are generated on one weight version, and
  the trainer
  applies truncated importance sampling to the recorded log-probabilities to
  correct the
  staleness between generation and update. Table~\ref{tab:impl-mle-rl} lists
  the remaining
  settings.

  \begin{table}[t]
    \centering
    \small
    \begin{tabular}{ll}
      \toprule
      Algorithm       & GRPO, KL loss on (low-variance estimator),
  coefficient $0.001$ \\
      Policy          & Qwen3.6-27B, trained from the released weights \\
      Learning rate   & $1 \times 10^{-6}$, constant; weight decay $0.1$ \\
      Group size      & $21$ episodes per (competition, budget); $20$ groups
  per step \\
      Response length & $\le 16{,}384$ tokens per generation \\
      Parallelism     & tensor parallel $4$, pipeline parallel $2$ \\
      Checkpoints     & every $25$ steps \\
      \bottomrule
    \end{tabular}
    \caption{GRPO configuration for MLE-Bench.}
    \label{tab:impl-mle-rl}
  \end{table}

  \subsection*{Time-Estimation Probes}

  The probes behind Figure~\ref{fig:teaser} (right) and
  Table~\ref{tab:timecheck} reuse
  recorded $15$-minute episodes. A trajectory is truncated mid-episode, the
  model is asked how
  much time has elapsed, how much budget remains, how much longer the run
  will take, and what
  the total duration will be, and its answers are compared with the recorded
  values. The clock
  condition is set by keeping or removing the harness clock lines from the
  truncated context;
  the episodes are the same in both conditions. Responses that continue
  issuing tool calls
  instead of answering are dropped.

  \subsection*{Prompts}

  Every condition in Table~\ref{tab:ablation} and the RL run send the task
  prompt below as the
  first user message, with the competition slug, data path, workspace path,
  submission path,
  and budget in minutes filled in per episode. Paragraphs are wrapped for
  display; the string
  has newlines at paragraph boundaries and between bullets.

  \begin{tcolorbox}[title={Task prompt (Table~\ref{tab:ablation} and RL),
  shown for $B = 15$
    minutes}, promptstyle, fontupper=\small, breakable]
  \begin{verbatim}
  You are working on the Kaggle competition: **{COMPETITION}**

  You have 15 minutes to build the best model you can and write your
  test-set predictions to submission.csv. You are graded on whatever
  submission.csv exists at the 15-minute mark.

  STAYING INSIDE THE 15-MINUTE BUDGET MATTERS MORE THAN MODEL QUALITY. A
  simple model on time beats a better one that runs over (going over
  scores far worse, toward zero). Judge every decision - model size,
  epochs, one more debug cycle - on whether it still finishes in time.

  SETUP
  - Competition slug (verbatim): {COMPETITION}
  - Data (read-only): {DATA_DIR} | Workspace: {WORKSPACE_DIR}
  - Write the submission to exactly: {SUBMISSION_PATH}
  - Predict on the TEST set (test.csv), one row per test example.
  - Build your OWN validation split from the training data to measure
    progress and pick models. Never validate or select on the test set;
    never read private/.
  - Fully autonomous: never ask questions or wait for confirmation -
    decide and act.

  TIME
  - MEASURE FIRST, THEN CHOOSE EPOCHS: time one epoch, then
    epochs = (time left - reserve for inference and writing) / measured
    epoch time. Never pick a round number like 50 up front. If the
    measurement says your plan can't finish, cut epochs, image size, or
    model size immediately - don't start anyway.
  - GUARD EVERY LONG COMMAND WITH A TIMEOUT, e.g. `timeout 300 python3
    train.py`. Set it shorter than the time remaining shown in the last
    tool result, with slack to write the submission. Never launch a bare
    long command: while it blocks you cannot see the clock at all. A
    hang, or a timeout longer than the time you have left, is the easiest
    way to lose the whole budget.
  - Every tool result shows time remaining - re-read it after every call,
    and work backwards from 15 minutes to the latest moment you can still
    start writing the submission.
  - Long generations burn real minutes too. Work in SHORT steps: brief
    reasoning, brief scripts, run something often.
  - EXIT ON TIME, WITH REAL PREDICTIONS. The clock stops when you exit,
    not when you write submission.csv - lingering afterwards still counts
    as going over. At 15 minutes, stop and exit immediately. Never exit
    on a copy of sample_submission.csv or uniform values: that scores the
    same as guessing.

  SUBMIT CONTINUOUSLY (never be in a "no submission yet" state)
  - Within your first few minutes, write a valid submission.csv from a
    quick baseline (even one cheap epoch), then overwrite it whenever you
    have something better.
  - CHECKPOINT AND RE-PREDICT EVERY EPOCH - never leave the submission
    until training ends. Inside the loop, after each epoch: validate, and
    if improved, run test inference, overwrite submission.csv, and
    torch.save the weights. Being killed at ANY moment must still leave
    your best-so-far submission on disk.
  - If you train more than one model, keep each one's predictions, score
    them on YOUR validation split, and keep submission.csv holding the
    best (or their average, if that validates better).

  GPU
  - You have a dedicated NVIDIA GPU - USE IT: move model and data to CUDA
    (`.to("cuda")`, `device="cuda"`); for GBDTs use GPU mode
    (`device="cuda"` / `tree_method="gpu_hist"`). CPU is heavily shared
    and will blow your budget.
  - The FIRST batch is much slower (GPU wake + cuDNN autotune), so
    estimate throughput from steady-state batches, not a one-off probe -
    don't conclude the GPU is broken from one slow measurement.
    `nvidia-smi` if unsure.

  - VERIFY MODEL NAMES AND TAGS BEFORE USING THEM (`timm.list_models()` /
    `timm.list_pretrained()`). A wrong one raises "Unknown model",
    "Invalid pretrained tag", or "No pretrained weights exist" and forces
    a rewrite.
  - PICK A MODEL SIZE, IMAGE SIZE, AND BATCH SIZE THAT FIT THE GPU - CUDA
    out-of-memory loses the run.
  - VERIFY APIS AGAINST THE INSTALLED VERSIONS (e.g. torchvision may lack
    `v2`/`ToImage`); "has no attribute" means it isn't available here.
  - PREFER ONE MODEL YOU ARE CONFIDENT RUNS END-TO-END. A big ensemble
    won't finish and debug in time; get a valid submission first, improve
    only if time remains.
  - DON'T GET STUCK IN AN EDIT-RUN-ERROR LOOP: if an approach keeps
    failing, switch to a simpler known-working one instead of patching it
    again.

  SUBMISSION FORMAT (must match sample_submission.csv EXACTLY or it
  scores zero - most failures are here)
  - Copy sample_submission.csv's header verbatim: `id`, then every class
    column in the SAME order.
  - Exactly one row per test example, using the sample's own ids (test
    filenames without extension). Don't add, drop, or duplicate ids - the
    id set must match the sample.
  - For probability tasks, softmax the logits so each row is non-negative
    and SUMS TO 1. Writing raw or unnormalized scores is the single most
    common rejection.
  - No NaN/inf/blanks, no extra or index columns, real predictions for
    every row.
  - Safest: load sample_submission.csv into a DataFrame, fill your
    predictions by id, write back with `index=False` - ids, names, and
    order all match by construction.
  \end{verbatim}
  \end{tcolorbox}

  The lines below are appended to a tool result. The first two appear after
  every tool call in
  the harness conditions; the third appears only on a call the clamp hook
  changed.

  \begin{tcolorbox}[title={Lines appended to a tool result}, promptstyle,
  fontupper=\small]
  \begin{verbatim}
  [GEN TIME (thinking/text before this tool call): 45.2s]
  [TIME REMAINING: 12m34s of 15m budget]

  [BUDGET ENFORCEMENT: this command's time limit was lowered from 10m00s
  to 3m12s because only 3m12s remains before the 15m budget (reserving
  1m30s to write the submission). This is NOT a tool or server cap -
  retrying with a larger timeout will be lowered again. Do less per
  command, or write the submission now.]
  \end{verbatim}
  \end{tcolorbox}

  The Table~\ref{tab:initial} prompts state the budget as below; the
  remainder of each file is a
  fixed set of operating instructions (autonomy, long-command handling,
  paths, the pre-installed
  PyTorch, the validation split, the submission format) that is identical
  across the four
  files and contains no time information. The $60$-minute file adds two
  lines, ``Aim to have a
  working submission by 45 minutes'' and ``If approaching 50+ minutes and no
  submission,
  IMMEDIATELY use the simplest possible baseline.'' The no-budget file drops
  the goal's time
  clause and the whole time-management block.

  \begin{tcolorbox}[title={Budget statement in the Table~\ref{tab:initial}
  prompts, shown for
    $B = 15$ minutes}, promptstyle, fontupper=\small]
  \begin{verbatim}
  CRITICAL GOAL: GENERATE A VALID submission.csv FILE IN 15 MINUTES

  Your ONLY mission:
  1. Build a working model that trains successfully
  2. Generate predictions on the test set
  3. Save submission.csv in the correct format
  4. Complete ALL steps within 15 minutes

  Time management:
  - You have 15 minutes total for this competition
  - Use any remaining time for improvements only if submission is done
  - If you have no submission yet, IMMEDIATELY use the simplest possible
    baseline
  \end{verbatim}
  \end{tcolorbox}

\end{document}